\documentclass[runningheads]{llncs}
\usepackage{eccv}
\usepackage{eccvabbrv}
\usepackage{hyperref}
\usepackage{url}
\usepackage{algorithm2e}
\usepackage{algpseudocode}
\SetKwInOut{Parameter}{parameter}
\usepackage{amsmath}  
\usepackage{graphicx}
\usepackage{amsmath}
\usepackage{amssymb}
\usepackage{booktabs}
\usepackage{colortbl}
\usepackage{caption}
\usepackage{subcaption}
\usepackage{csquotes}
\usepackage{pifont}
\usepackage{color,soul}
\usepackage{booktabs}
\usepackage{subcaption}
\usepackage{multirow}
\usepackage{float}
\usepackage{array}
\usepackage{pifont}
\usepackage{xcolor} 
\usepackage{makecell} 
\usepackage{graphicx}    
\usepackage{subcaption} 
\usepackage{csquotes}
\usepackage{listings}
\usepackage{caption}
\usepackage{subcaption}
\usepackage{booktabs}
\usepackage{hyperref}
\usepackage{adjustbox}
\definecolor{lightblue}{RGB}{230, 242, 255}
\definecolor{lightgray}{gray}{0.95}
\definecolor{sectionred}{RGB}{255, 230, 230}   
\definecolor{rowbluegray}{RGB}{235, 240, 250}  
\usepackage{graphicx}
\usepackage{booktabs}

\usepackage[accsupp]{axessibility}  
\usepackage{hyperref}

\usepackage{orcidlink}

\begin{document}

\title{ SpEmoC: A Balanced Speaker-Segment Multimodal Emotion Benchmark} 

\titlerunning{SpEmoC: A Balanced Speaker-Segment Multimodal Emotion Benchmark}

\author{Sania Bano\inst{1}\orcidlink{0009-0009-3134-108X} 
 \and
Shahzad Ahmad\inst{2}\orcidlink{0009-0002-5962-2659} \and
Santosh Kumar Vipparthi\inst{1}\orcidlink{0000-0002-5672-3537}\and
Sukalpa Chanda\inst{2} \orcidlink{0000-0002-9068-5845
}\and
Subrahmanyam Murala\inst{3}\orcidlink{0000-0003-3384-4368} }
\authorrunning{S.~Bano et al.}

\institute{CVPR Lab, Indian Institute of Technology Ropar, India\and
Østfold University College, Norway \\
\and
CVPR Lab, SCSS, Trinity College Dublin, Ireland\\
\email{sania.22eez0012@iitrpr.ac.in}}

\maketitle

\begin{figure}[h]
    \centering
    \includegraphics[width=\textwidth]{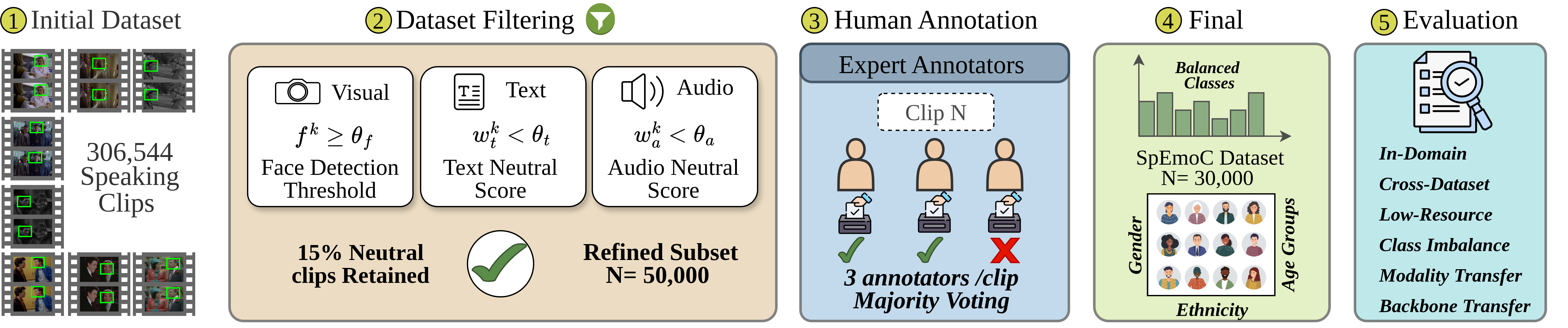}
\caption{Overview of SpEmoC construction and evaluation. Multimodal filtering and expert validation refine 306K speaking clips into 30K balanced clips across seven emotions for diverse benchmarking.}

    \label{filter_human_anno}
    \vspace{-10mm}
\end{figure}

\begin{abstract}
Understanding human emotions in spoken conversations is a key challenge in affective computing, with applications in empathetic AI, human computer interaction, and mental health monitoring. However, existing datasets vary in scale, emotion distribution, modality alignment, and data partitioning strategies, which can influence reliable cross-dataset generalization and minority-emotion modeling. We introduce \textbf{SpEmoC} a \textbf{Sp}eaking segment \textbf{Emo}tion for \textbf{C}onversations comprising 306,544 raw clips from 3,100 English language movies and TV series. From these, 30,000 high quality, class balanced clips are curated, featuring synchronized visual, audio, and textual modalities annotated for seven emotions through a hybrid pipeline that integrates pretrained models with human validation. SpEmoC uses strict movie- and series-level splits to prevent content overlap between split sets, allowing more reliable evaluation of model generalization. The dataset also maintains a near-balanced distribution across seven emotions, including minority classes such as \textit{Fear} and \textit{Disgust}, which supports more balanced learning across categories. Extensive experiments, including in-domain benchmarking, cross-dataset transfer, low-data training, class-imbalance analysis, and modality transfer show that balanced data and careful splitting lead to more stable performance across emotions when models are evaluated on other datasets. These results highlight the importance of dataset design for robust and transferable multimodal emotion recognition.\\

\noindent\textbf{Project Page:} The full dataset is available at the following  \href{https://skvipparthi.com/spemoc.html}{link}

\end{abstract}
\vspace{-4mm}

\section{Introduction}
\vspace{-2mm}
Understanding emotions in spoken conversations is central to affective computing, with applications in empathetic AI, mental health monitoring, pain detection~\cite{lucey2011pain}, and human computer interaction~\cite{ picard1997affective}. Emotion is inherently multimodal expressed through facial expressions, vocal prosody, and linguistic context yet robust multimodal modeling in natural conversations remains a major challenge. While modern multimodal models have advanced rapidly, robust emotion recognition in natural conversational settings remains fundamentally constrained by benchmark design. The structure of the dataset, including how clips are segmented, class distribution, and the partitioning strategy, directly influences generalization stability and performance on minority emotions.

\begin{figure*}[htbp!]
\centering
\begin{subfigure}[b]{0.52\textwidth}
    \centering
    \includegraphics[width=\textwidth]{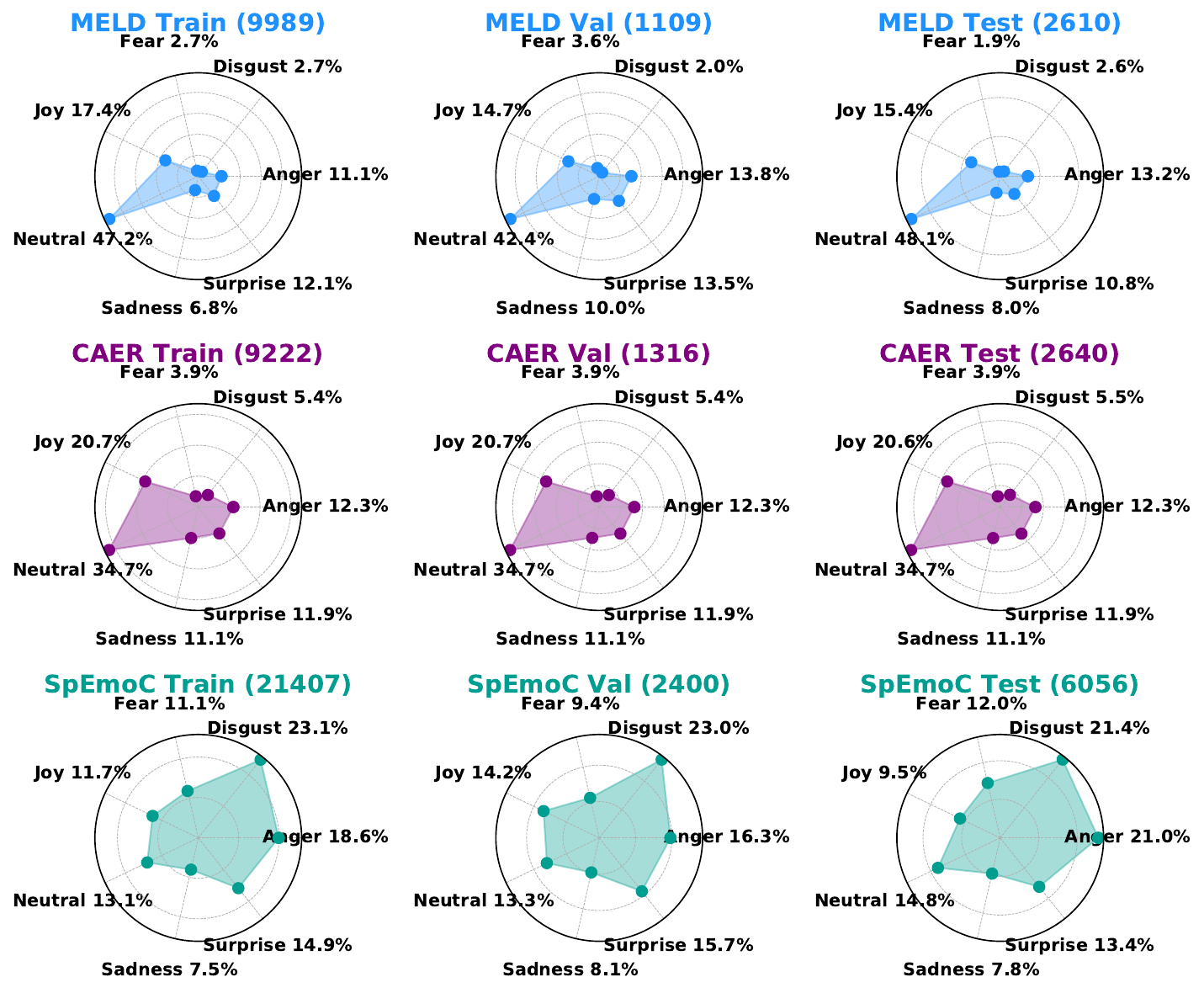}
    \caption{}
    \label{subfig:class_dist}
\end{subfigure}
\hfill
\begin{subfigure}[b]{0.44\textwidth}
    \centering
    \includegraphics[width=\textwidth]{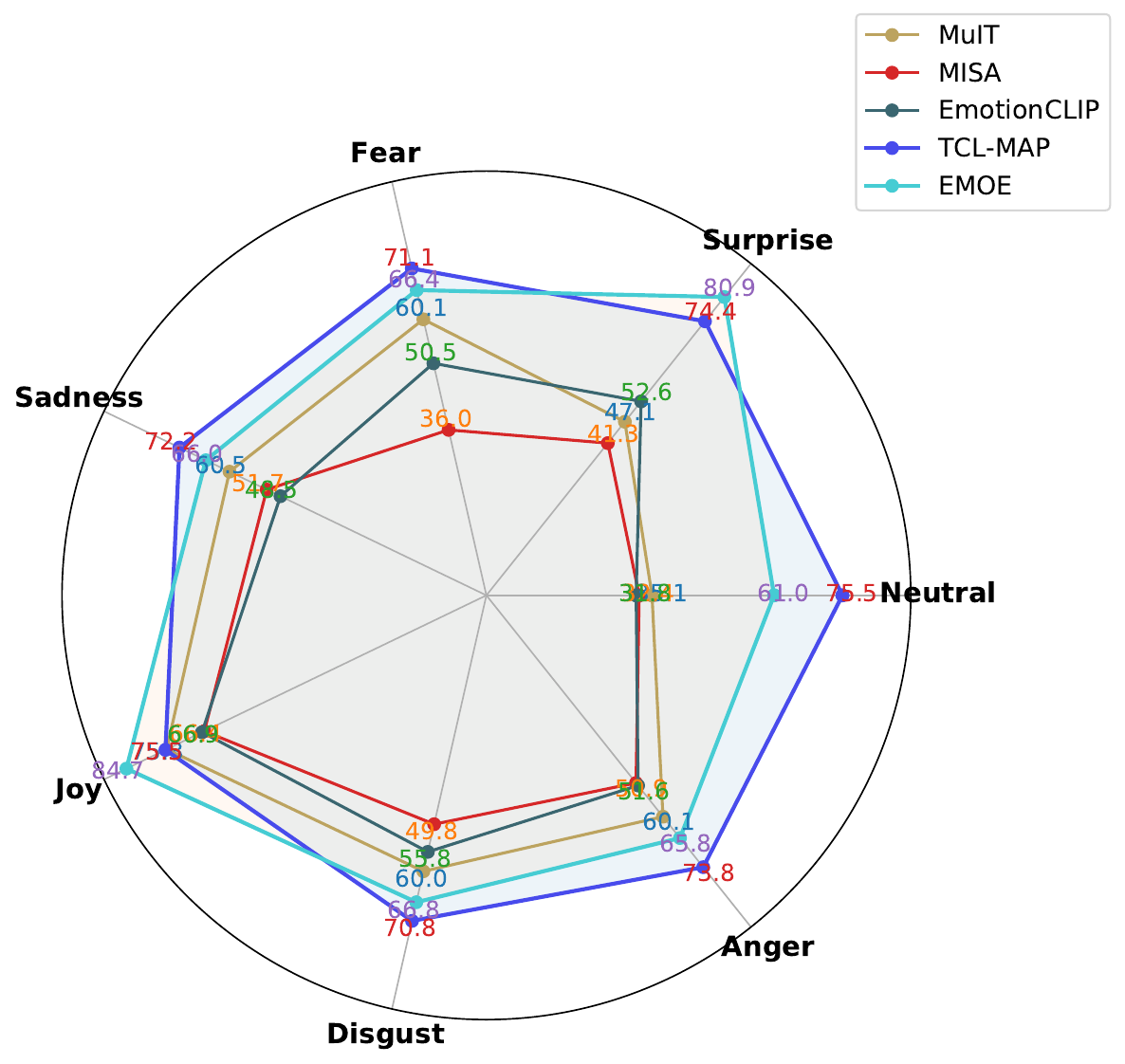}
    \caption{}
    \label{subfig:perf_comp}
\end{subfigure}
\vspace{-2mm}
\caption{(a) Emotion class distribution across train, validation, and test splits for \textbf{SpEmoC}, MELD~\cite{poria2019meld}, and CAER~\cite{lee2019context}. SpEmoC exhibits a more balanced distribution across all seven emotions, while MELD and CAER are dominated by the Neutral class. (b) Class-wise baseline performance on SpEmoC, showing more consistent F1-scores due to the balanced emotion distribution.}
\label{distribution}
\vspace{-10pt}
\end{figure*}

Multimodal Emotion Recognition (MER) faces significant challenges that hinder its deployment in dynamic, dialogue-driven contexts. Most existing benchmarks focus on unimodal settings, such as facial expressions \cite{li2017reliable, zeng2018facial} or audio \cite{schuller2011recognising, el2011survey}. They suffer from limited modality alignment and annotation scale \cite{poria2017review, zadeh2018multimodal, albanie2018emotion}. 
Datasets like the Multimodal EmotionLines Dataset (MELD) \cite{poria2019meld}, with 13,000 utterances from the TV series \textit{Friends}, and CAER \cite{lee2019context}, which includes 13,201 video clips from 79 TV shows with audio and visual tracks, offer multimodal annotations at limited scales. Similarly, EmoWOZ \cite{feng2022emowoz} provides 11,000+ task-oriented dialogue utterances with multimodal labels. However, these datasets are constrained by relatively small sizes and imbalanced emotion distributions, with MELD and CAER dominated by "Neutral" emotions and underrepresentation of "Fear" and "Disgust" (see Figure \ref{distribution}(a)). These datasets lack real-world diversity and, being built from TV series, often reuse characters across splits, making test sets not truly unseen. Similarly, the M3ED dataset \cite{zhao2022m3ed} offers 9,000 utterances from TV series but falls short in capturing the breadth of emotional expressions needed for generalization. Furthermore, the reliance on expensive human annotations and the absence of synchronized multimodal alignment limit the scalability and applicability of these datasets. Fusing heterogeneous modalities also remains challenging due to differences in data representation, temporal dynamics, and emotional relevance across text, audio, and visual streams \cite{zadeh2018multimodal, 6487473}.\par

To address these limitations, we introduce \textbf{SpEmoC}, a large-scale \textbf{Sp}eaking-segment \textbf{Emo}tion for \textbf{C}onversations dataset constructed from diverse movies and TV series. SpEmoC provides synchronized video, audio, and text modalities, focusing on natural speaking segments that convey explicit emotional intent under varied real-world conditions. A hybrid annotation framework integrates pretrained text and speech models with human validation to ensure accurate and balanced labeling across seven emotion categories.\\
To comprehensively evaluate the dataset, we benchmark several state-of-the-art multimodal emotion recognition models on SpEmoC and conduct extensive cross-dataset evaluations on MELD and CAER. These experiments demonstrate that balanced supervision yields more stable and transferable performance across emotion categories, particularly for previously underrepresented classes.
In addition, we perform detailed ablation studies including low-data regimes, cross-dataset fine-tuning, neutral-class removal, and unimodal versus multimodal analysis to validate the robustness and practical utility of SpEmoC for real-world multimodal emotion understanding. The contributions of the paper are as follows: 
\begin{itemize}
    \item A large scale, class balanced dataset of speaking segment clips with tightly aligned visual, auditory, and linguistic modalities.
    \item  A scalable multimodal annotation pipeline integrating pretrained text and speech models with human validation to ensure synchronized alignment and high-quality emotion labeling.    
   \item Through cross-dataset evaluations on MELD and CAER with multiple SOTA models, we show that SpEmoC enables more balanced class-wise performance and improved robustness to class imbalance, particularly for minority emotions.
    \item We provide a comprehensive empirical study of multimodal generalization, including low-resource settings, neutral-class analysis, and unimodal versus multimodal comparisons, offering insights into benchmark design for robust affective modeling.

\end{itemize}
SpEmoC advances multimodal emotion recognition through larger scale, balanced classes, and speaking-segment alignment, providing a stable and transferable benchmark for affect modeling.
 
\vspace{-4mm}
\section{Related Work}
\vspace{-3mm}
\textbf{Emotion Datasets:}
Early affective computing studies relied on acted or studio-controlled corpora such as IEMOCAP~\cite{busso2008iemocap}, RAVDESS \cite{livingstone2018ryerson}, and EmoReact~\cite{nojavanasghari2016emoreact}, which offered limited expressive diversity and small emotional vocabularies. 
Later multimodal datasets integrated video, audio, and text for conversational emotion analysis. 
MELD~\cite{poria2019meld} and CAER~\cite{lee2019context} expanded to TV-series dialogues but remain small ($<14$k samples) and heavily skewed toward neutral expressions. 
CMU-MOSEI~\cite{zadeh2018multimodal}  covers monologue-based content, while M3ED~\cite{zhao2022m3ed}  and EmoWOZ~\cite{feng2022emowoz} provide task-oriented or synthetic dialogues. 
Recent large-scale efforts such as EmotionTalk~\cite{sun2026emotiontalk}, EMOVOME~\cite{gomez2024emovome}, and EmotionCLIP~\cite{zhang2023learning} improve coverage through weak supervision or multilingual expansion but still lack balanced, human-validated, temporally aligned annotations. EmotionLLAMA’s MERR \cite{cheng2024emotion} includes 28,618 coarse- and 4,487 fine-grained samples, many from non-English dubbed videos.
A concise comparison in Table~\ref{tab:multimodal_comparison} highlights these limitations and motivates the design of \textbf{SpEmoC} as a fair, large-scale, and culturally diverse benchmark.

\noindent\textbf{Model-Level Advances:}
Multimodal emotion recognition models have evolved from early feature fusion methods to attention-based and contrastive learning frameworks. Transformer-based architectures such as MulT~\cite{tsai2019multimodal}  and MISA~\cite{hazarika2020misa} model cross-modal interactions between text, audio, and visual streams, while recent approaches including TCL-MAP~\cite{zhou2024token} and EMOE~\cite{fang2025emoe} further improve multimodal fusion and representation learning. EmotionCLIP~\cite{zhang2023learning} leverages vision-language contrastive pretraining to align emotional semantics across modalities. Despite these advances, model performance remains strongly dependent on the quality and balance of available datasets, which motivates the development of more robust multimodal benchmarks such as SpEmoC.
\vspace{-4mm}

\begin{table*}[htbp!]
\centering
\caption{Comparison with existing multimodal emotion recognition datasets. Modalities: A = Audio, V = Visual, T = Text.}
\vspace{-2mm}
\scriptsize
\label{tab:multimodal_comparison}
\renewcommand{\arraystretch}{0.5}
\setlength{\tabcolsep}{5pt}
\begin{tabular}{p{3cm} c c c c p{2.8cm}}
\toprule
\textbf{Dataset} & \textbf{\#Clips} & \textbf{Modalities} & \textbf{\#Emotions} & \textbf{Source} \\
\midrule

IEMOCAP~\cite{busso2008iemocap} & 10k & V, A, T & 6 & Acted dialogues \\

CMU-MOSEI~\cite{zadeh2018multimodal} & 23k & V, A, T & 6 & YouTube \\

MELD~\cite{poria2019meld} & 13k & V, A, T & 7 & Friends TV show \\

CAER~\cite{lee2019context} & 13k & V, A & 7 & TV shows \\

RAVDESS~\cite{livingstone2018ryerson} & 7k & V, A & 8 & Studio-acted clips \\

EmoReact~\cite{nojavanasghari2016emoreact} & 1k & V, A & 8 & YouTube videos \\

\midrule
\rowcolor{lightblue}
\textbf{SpEmoC (Ours)} & \makecell{\textbf{306,544} (raw),\\\textbf{30k} (refined)} & \textbf{V, A, T} & \textbf{7} & \textbf{Movies \& TV series} \\

\bottomrule
\end{tabular}
\end{table*}

\vspace{-7mm}
\section{SpEmoC Dataset Construction}
\label{sec:dataset}
\begin{figure}[htbp!]
    \centering
    \includegraphics[width=.8\textwidth]{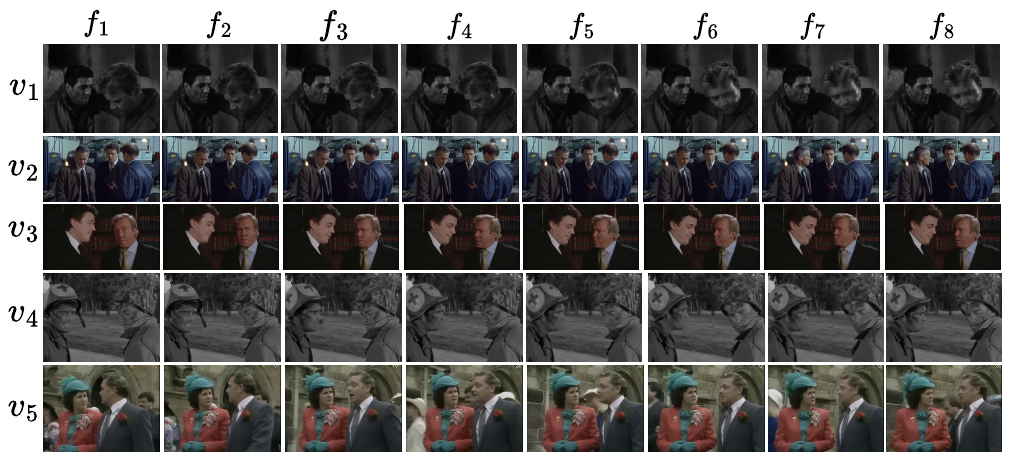}
    \vspace{-2mm}
    \caption{Examples from SpEmoC showing variation in genre, lighting, color, and expression. Each row displays 8 sampled frames from a distinct clip.}
    \label{fig:spemoc_examples_colg}
    \vspace{-3pt}
\end{figure}

\vspace{-3mm}
Existing multimodal emotion datasets are often limited in scale, synchronization, and emotional diversity, which restricts the development of robust and generalizable models. Many datasets rely on short caption-based clips or controlled acted expressions and fail to capture natural emotional interactions. They also exhibit strong class imbalance, with the \textit{neutral} class dominating and minority emotions such as \textit{fear} and \textit{disgust} underrepresented. These limitations highlight the need for a larger, balanced dataset that better reflects emotional variability across practical conversational scenarios.\\
To address these gaps, we introduce \textbf{SpEmoC}, a large-scale multimodal emotion recognition dataset focused on \textbf{speaking segments} with synchronized \textbf{video}, \textbf{audio}, and \textbf{text}. SpEmoC contains 30,000 refined clips curated from 306,544 raw segments collected from 3,100 English-language movies and TV series selected for their emotionally expressive dialogues. The dataset captures diverse genres (e.g., drama, comedy, horror, romance, thriller), formats (color and black-and-white), and environmental conditions, reflecting realistic emotional expressions with temporally aligned multimodal signals. Each segment preserves authentic speech, facial expressions, and contextual dialogue while excluding dubbed or subtitled content to maintain linguistic fidelity. Through targeted filtering and refinement, SpEmoC achieves a balanced distribution across seven emotion classes (Figure \ref{distribution}). Table~\ref{tab:dataset_overview} summarizes the dataset characteristics, while Figure \ref{fig:spemoc_examples_colg} illustrates its visual and emotional diversity. Additional details of the data curation pipeline are provided in the \textcolor{blue}{\textbf{\textit{Supplementary Sec: 4}}}.
\vspace{-5mm}
\vspace{-6mm}
\subsection{Data Collection and Processing Pipeline}
\vspace{-2mm}
We develop a scalable multi-stage pipeline that converts long-form videos into synchronized \textbf{video}, \textbf{audio}, and \textbf{text} segments (Figure \ref{fig:spemoc_pipeline}), ensuring temporal alignment and modality consistency. Let \( \mathcal{V} = \{V_1, V_2, \dots, V_{3100}\} \) be the set of source videos, each \( V_k \) with time duration \(\geq 40 \) minutes.  The stages are outlined below.\\
\textbf{a. Dialogue Segmentation:}
We use the Whisper ASR model~\cite{radford2023robust} to transcribe each video with word-level timestamps. A video clip segment \( v_{k}^{i} \)(\(i^{th}\) clip of video \(V_k\))  is retained if it (i) contains at least 12 words, ensuring sufficient linguistic and emotional context, and (ii) ends with terminal punctuation (\texttt{.}, \texttt{!}, \texttt{?}) to capture a complete utterance.
Each video \( V_k \) yields  a set of segments \( \{v_{k}^{1}, v_{k}^{2}, \dots, v_{k}^{m}\} \). The total number of extracted clips is:
\vspace{-1.5mm}
\begin{equation}
\scriptsize
    N_{\text{clips}} = \sum_{k=1}^{N} \sum_{i=1}^{m} 1 \approx 306{,}544
\end{equation}
where \( m \) is the number of valid clip segments in \( V_k \) and \(N = |\mathcal{V}|\). This segmentation ensures meaningful dialogue boundaries rather than arbitrary frame windows.\\
The extracted clips are short dialogue segments (typically 3-6 seconds) and therefore usually contain a single dominant speaker. During the human verification stage, annotators review the visual, audio, and textual modalities jointly to determine the emotion expressed in the clip. We do not explicitly perform speaker diarization or active speaker detection; instead, the assigned label corresponds to the dominant emotional expression within the speaking segment.

\noindent\textbf{b. Multimodal Extraction:} After segmentation, each valid clip \( v_{k}^{i} \) is decomposed into three aligned modalities:
\vspace{-2mm}
\begin{itemize}
  \item \textbf{Visual:} Video segment \( v_{k}^{i} \) from the same temporal window, containing number of frames 
  \( |v_{k}^{i}| = (t_{end}-t_{start}) \times \text{FPS} \). This process yields approximately 30 million frames across all clips.
  \item \textbf{Text:} Transcript \( \mathcal{T}_{k}^{i} \) of the corresponding \( v_{k}^{i} \) clip, obtained from Whisper. 
  \item \textbf{Audio:} Waveform  \( \mathcal{A}_{k}^{i} \) of the corresponding \( v_{k}^{i} \) clip, cropped from the interval \([t_{\text{start}}, t_{\text{end}}]\), using FFmpeg (open-source toolkit) \cite{ffmpeg2025}, Where $t_{\text{start}}$ is start time and $t_{\text{end}}$ is end time.
\item \textbf{Human and Face Detection:} Human bodies and faces are detected in each frame of \(v_{k}^{i}\) using YOLOv8~\cite{varghese2024yolov8} to verify the presence of a visible person in the clip \(v_{k}^{i}\). 

\end{itemize}
\vspace{-1mm}
\noindent\textbf{c. Synchronization Check:}
Temporal alignment between modalities is verified by comparing audio and video durations (timestamp consistency):
\begin{equation}
\scriptsize
    |\text{Duration}(\mathcal{A}_{k}^{i}) - (t_{end}(v_{k}^{i}) - t_{start}(v_{k}^{i})| \leq \epsilon
\end{equation}

Here, \( \epsilon = 0.1~\text{sec} \) denotes the maximum allowable deviation. Segments violating this constraint are reprocessed or excluded.\\
Overall, this preprocessing pipeline, comprising segmentation, multimodal extraction, and synchronization followed by post-processing, produces a high-quality, temporally aligned clips for multimodal annotation and analysis.
\vspace{-3mm}
\begin{table*}[htbp!]
\centering
\scriptsize
\caption{Overview of the Proposed Multimodal Emotion Recognition SpEmoC Dataset, Highlighting Source, Structure, Modalities, Annotation, Splitting policy, and Emotion Coverage.}
\vspace{-3mm}
\label{tab:dataset_overview}
\renewcommand{\arraystretch}{1}
\begin{tabular}{p{4.5cm} p{7.7cm}}
\rowcolor{yellow!22} \multicolumn{2}{l}{\textbf{Data Source and Composition}} \\
\rowcolor{lightblue} Source & YouTube (Movies and TV Series) \\
\rowcolor{lightblue}Number of Videos & 3,100 \\
\rowcolor{lightblue}Video Length & $\ge$ 40 minutes each \\
\rowcolor{lightblue}Video Types & Color and Black-and-White , English language , Non-dubbing, Subtitle independent \\
\rowcolor{lightblue}Video genres &  Drama, Comedy, Horror, Thriller, Romance, History, etc. \\
\rowcolor{lightblue}Total Number of Clips & 306,544 \\
\rowcolor{lightblue}Average Clip Duration & 3–6 seconds \\
\rowcolor{lightblue}Total Frames & 30 million+ \\
\rowcolor{lightblue}Focus Per Clip & Speaking segments only \\

\rowcolor{yellow!22} \multicolumn{2}{l}{\textbf{Modalities and Preprocessing}} \\
\rowcolor{lightblue} Modalities & Video, Audio, Text \\
\rowcolor{lightblue}Face/Human Detection & YOLOv8 \cite{varghese2024yolov8} \\
\rowcolor{lightblue}Face Presence Threshold & Face detected in $\ge$ 90\% of frames \\

\rowcolor{yellow!22} \multicolumn{2}{l}{\textbf{Annotation and Labeling Strategy}} \\
\rowcolor{lightblue} Annotation Models & DistilRoBERTa (Text) \cite{sanh2019distilbert}, Wav2Vec 2.0 (Audio) \cite{baevski2020wav2vec} \\
\rowcolor{lightblue}Label Type & Single dominant emotion per clip \\
\rowcolor{lightblue}Label Fusion & Logit-Based Fusion from Text and Audio modalities \\
\rowcolor{yellow!22} \multicolumn{2}{l}{\textbf{Dataset Spliting Strategy }} \\
\rowcolor{lightblue}Movie/franchise-level &All clips from the same movie, sequel, or multi-episode series are assigned exclusively to one split (train, validation, or test) to ensure a fully unseen test set. \\
\rowcolor{yellow!22} \multicolumn{2}{l}{\textbf{Emotion Classes}} \\
\rowcolor{lightblue} Categories & Anger, Disgust, Fear, Joy, Sadness, Surprise, Neutral \\
\end{tabular}
\end{table*}

\subsection{Annotation Methodology}
\vspace{-2mm}
We adopt a hybrid annotation pipeline that integrates automated multimodal labeling with human validation and dataset refinement to ensure both scalability and reliability. 
Pretrained emotion recognition models first generate automated annotations across text and audio modalities, followed by systematic filtering and expert validation to refine ambiguous or low-confidence samples. 
This design balances large-scale automation with human oversight, yielding a high-quality, balanced multimodal dataset.\\
\textbf{Emotion Taxonomy:}
Following Ekman’s basic-emotion taxonomy~\cite{ekman1992argument}, we adopt seven discrete categories for cross-dataset comparability:
E = \{\text{Anger}, \text{Disgust}, \text{Fear}, \text{Joy}, \text{Sadness}, \text{Surprise}, \text{Neutral}\}\\
Domain-specific pretrained models are used for text and audio, sharing identical emotion vocabularies to maintain cross-modal consistency.\\
\noindent\textbf{Automated Multimodal Annotation:}
Each dialogue transcript $(\mathcal{T}_{k}^{i})$ is analyzed using a fine-tuned DistilRoBERTa model~\cite{sanh2019distilbert}. For each utterance, we obtain a vector of real-valued scores called sentiment logits, representing the unnormalized model confidence for each emotion class in $E$.
\vspace{-2mm}
\begin{equation}
    \scriptsize
    {L_{t,}}_k^{i} = \text{logits}_{\text{text}}(\mathcal{T}_{k}^{i}) \in \mathbb{R}^{|E|}
\end{equation}
where  \({L_{t,}}_k^{i} = [{l_{t,}}_{k,1}^{i}, \dots, {l_{t,}}_{k,7}^{i}]\) contains model confidence scores for all emotion classes.
Similarly, the corresponding acoustic segment \(\mathcal{A}_{k}^{i}\) is processed using a pretrained Wav2Vec~2.0 model~\cite{baevski2020wav2vec}, yielding
\vspace{-2mm}
\begin{equation}
    \scriptsize
    {L_{a,}}_{k}^{i} = \text{logits}_{\text{audio}}(\mathcal{A}_{k}^{i}) \in \mathbb{R}^{|E|}
\end{equation}
where \({L_{a,}}_{k}^{i} = [{l_{a,}}_{k,1}^{i}, \dots, {l_{a,}}_{k,7}^{i}]\) represents the audio-domain logits.\\

\noindent\textbf{Logit-Based Multimodal Fusion for Emotion Labeling:}
\label{sec:fusion}
The sentiment logits \( L_{t} \) and \( L_{a}\) are used as inputs to our logit-based fusion mechanism to generate consistent pseudo-labels without relying on the noisy visual modality. Instead of using only top-class predictions, we leverage the full logit distributions to capture richer emotion signals and uncertainty across modalities.\\ 
For a fixed emotion label set \( E = \{e_1, \dots, e_7\} \), each clip produces two 7-dimensional logit vectors: \( L_t = [l_{t,1}, \dots, l_{t,7}] \) from the text model and \( L_a = [l_{a,1}, \dots, l_{a,7}] \) from the audio model, where the logits are unnormalized scores \( l_{m,i} \in \mathbb{R} \) with \( m \in \{t, a\} \).

Assuming a uniform prior \( P(e_i) = 1/|E| \), the posterior over emotion class \( e_i \) is modeled as \(P(e_i | L_t, L_a) \propto P(L_t | e_i) \cdot P(L_a | e_i)\)

where likelihoods are softmax-normalized:
\vspace{-2mm}
\begin{equation}
    \scriptsize
    \tilde{P}_m(e_i) = \frac{\exp(l_{m,i})}{\sum_{j=1}^{7} \exp(l_{m,j})}, \quad m \in \{t, a\}
\end{equation}

where \(l_{m,i}\) denotes the logit for class \(e_i\) from modality \( m \in \{t, a\} \).To encourage modality agreement, we introduce a KL-divergence penalty:
\vspace{-2mm}
\begin{equation}
    \scriptsize
    D_{\text{KL}}(\tilde{P}_t || \tilde{P}_a) = \sum_{i=1}^{7} \tilde{P}_t(e_i) \log \left( \frac{\tilde{P}_t(e_i)}{\tilde{P}_a(e_i)} \right)
\end{equation}

This yields a fused decision score:
\vspace{-2mm}
\begin{equation}
    \scriptsize
    S(e_i) = \log \tilde{P}_t(e_i) + \log \tilde{P}_a(e_i) - \lambda D_{\text{KL}}(\tilde{P}_t || \tilde{P}_a)
\end{equation}

where \(\lambda = 0.5\) controls the regularization strength. The predicted label and confidence are
\vspace{-2mm}
\begin{equation}
    \scriptsize
    e^* = \arg\max_{e_i \in E} S(e_i) , \quad F(e^*) = \frac{1}{1 + \exp(-S(e^*))}, \quad F(e^*) \in [0, 1]
\end{equation}
This formulation captures full distributional uncertainty from both modalities and enforces semantic coherence between textual and acoustic cues. Unlike majority voting or hard max fusion, the KL divergence regularizer penalizes disagreements and rewards confident, aligned predictions.

\begin{figure}[htbp!]
    \centering
    \includegraphics[width=\textwidth]{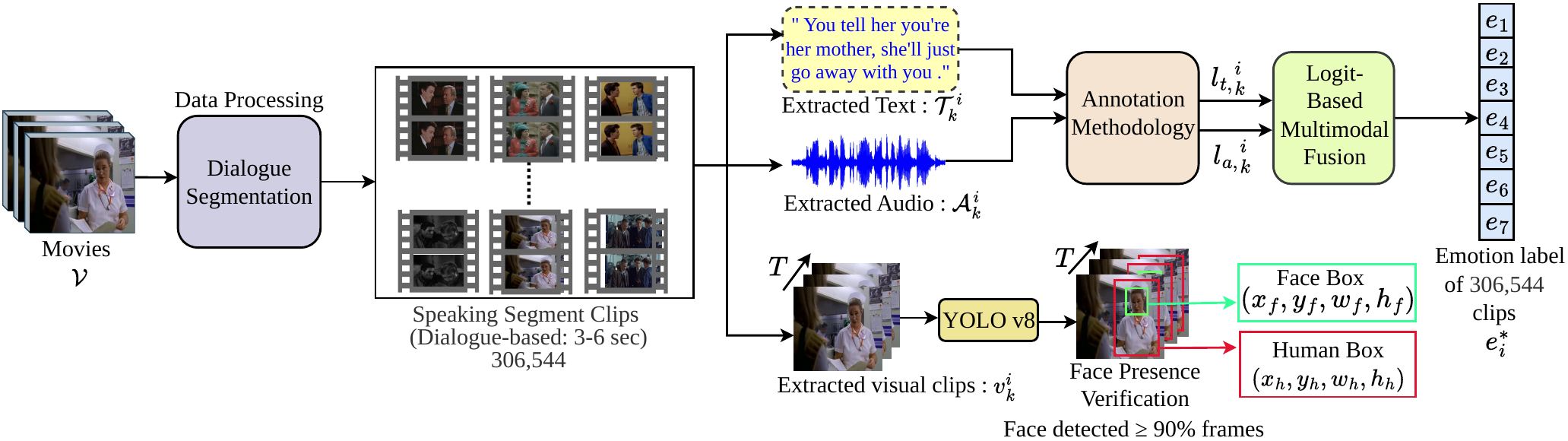}
    \vspace{-3mm}
    \caption{SpEmoC construction and multimodal labeling pipeline. Movies are segmented into short dialogue-based clips (3-6 s), from which synchronized text, audio, and visual streams are extracted. Text and audio logits are fused to obtain clip-level emotion labels, while YOLOv8 verifies face presence in visual frames ($\geq 90\%$ frames).}
    \label{fig:spemoc_pipeline}
    \vspace{-10mm}
\end{figure}

We further use the agreement condition \( \arg\max L_t = \arg\max L_a \) and confidence score \( F(e^*) \) to filter noisy labels.
This approach scales efficiently across large unlabeled datasets and acts as a pseudo-supervisor for high-quality emotion annotation.
Figure \ref{fig:spemoc_pipeline} illustrates the 
automated construction and multimodal labeling pipeline 
of the SpEmoC dataset. Further information on the annotation file is provided in the \textcolor{blue}{\textbf{\textit{Supplementary Sec: 5}}}.

\subsection{Dataset Refinement and Filtering}

\begin{figure*}[htbp!]
    \centering   
    \includegraphics[width=0.45\textwidth]{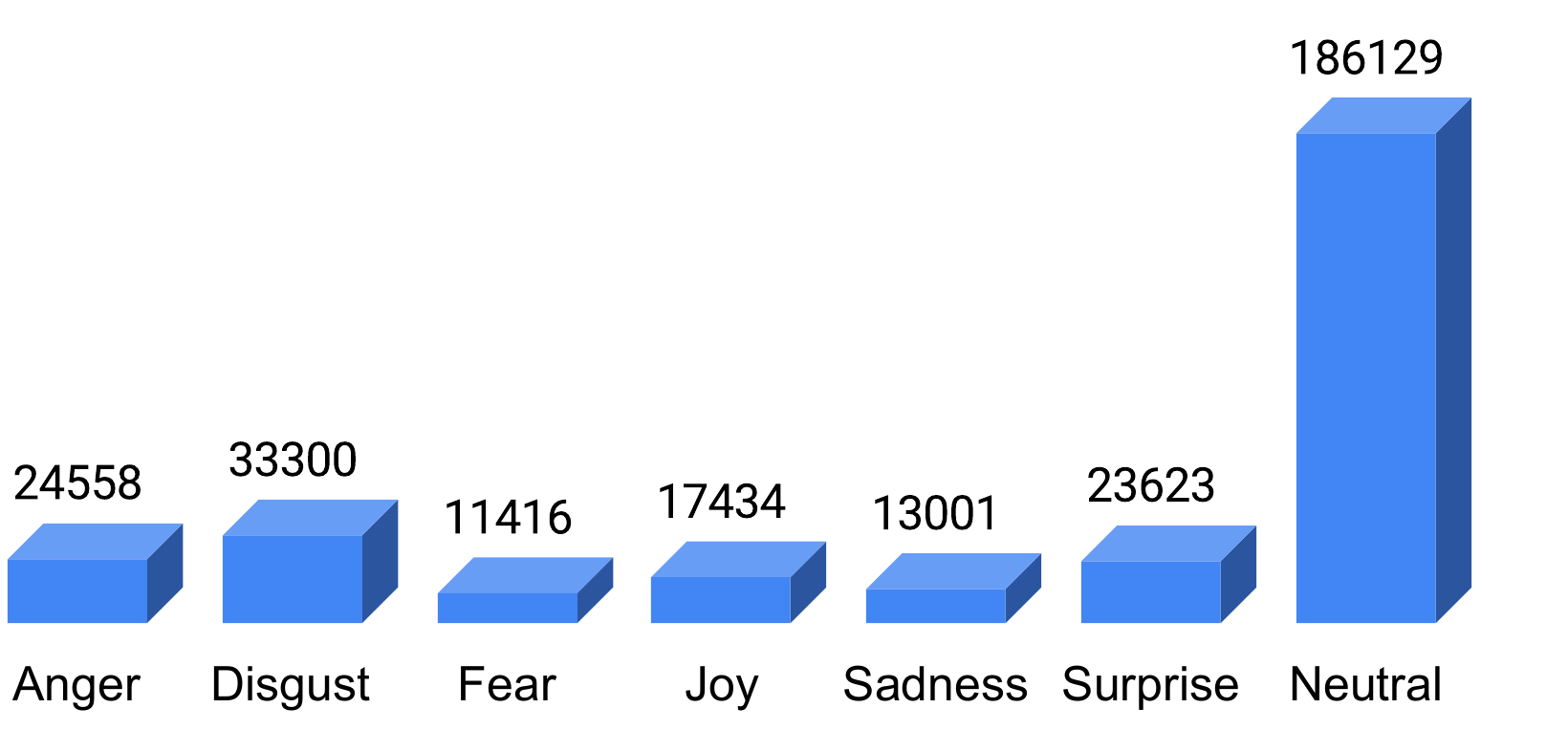}
    \includegraphics[width=.45\textwidth]{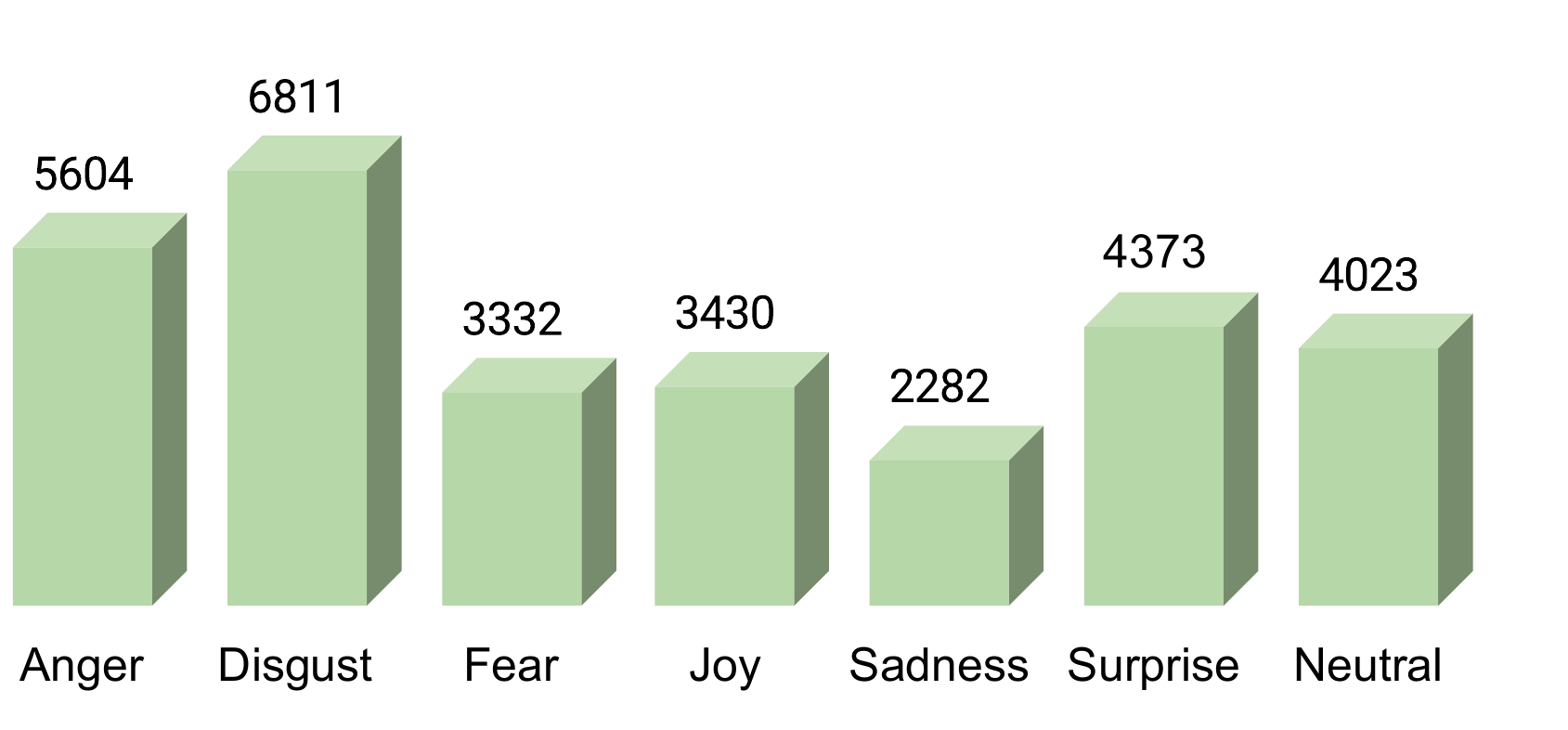}
\caption{Emotion class distribution before (blue) and after filtering (green). The initial 306K clips were dominated by the \textit{neutral} class, while \textit{fear} was underrepresented. A two-step filtering process was applied: threshold-based filtering, followed by human validation to remove ambiguous samples. The resulting 30,000-clip dataset achieves a more balanced distribution across all seven emotion classes.}
\small
\label{filter_dataset}
\vspace{-5mm}
\end{figure*}

The initial 306,544 annotated clips exhibited a strong class imbalance dominated by the neutral class (Figure~\ref{filter_dataset}), potentially biasing model training. To address this, we applied a multi-step filtering strategy using threshold-based constraints on text and audio neutral scores, along with face-presence verification in visual frames, to retain emotionally expressive clips. Thresholds were determined through manual experimentation and validated via performance analysis, revealing that many neutral clips contained weak emotional content reflected by high neutral scores. Figure~\ref{filter_human_anno} illustrates the complete dataset refinement and human annotation workflow.
For a clip \(v_{k}^{i}\),
\vspace{-2mm}
\begin{itemize}
    \item \({L_{t,}}_{k}^{i} = \text{text\_neutral\_logit}(\mathcal{T}_{k}^{i})\): text neutrality logit
    \item \({L_{a},}_k^i = \text{audio\_neutral\_logit}(\mathcal{A}_k^{i})\): audio neutrality logit
    \item \(f_k^{i} = \text{has\_face}(v_k^{i}) \in [0,1]\): face-presence confidence in video frames
\end{itemize}
\vspace{-2mm}
Logits are converted to probabilities via the sigmoid function:
\vspace{-2mm}
\[
\scriptsize
{w_{t,}}_k^{i} = \sigma({L_{t,}}_{k}^{i}) = \frac{1}{1 + e^{-{{L_{t,}}_{k}^{i}}}}, \quad
{w_{a,}}_k^i = \sigma({L_{a},}_k^i) = \frac{1}{1 + e^{-{{L_{a},}_k^i}}}
\]
A clip \(v_k^{i}\) is retained if: $f_k^i \ge \theta_f, \quad {w_{t,}}_k^i < \theta_t, \quad {w_{a,}}_k^i < \theta_a$ \\
where \(\theta_f = 0.9\) face presence in at least 90\% of frames, and \(\theta_t = \theta_a = 0.05\) exclude weakly emotional (neutral) segments.
A small portion of neutral clips (\(\sim15\%\)) is retained for balance.
This filtering reduced the pool to 50,000 high-quality clips with a significantly flatter class distribution.

\noindent\textbf{Human Validation of Labels:}
Following automatic filtering, 50,000 clips were subjected to human verification.
Twenty expert annotators, proficient in English and trained under standardized Ekman-based guidelines~\cite{ekman1992argument}, evaluated each clip using all three modalities (text, audio, visual).
Each sample was independently annotated by at least three experts, and final labels were determined through majority voting. Annotators did not have access to model-predicted labels during this process.
Inter-annotator agreement reached Fleiss’ \(\kappa = 0.62\) (\textit{substantial agreement}~\cite{landis1977measurement}).
Ambiguous or inconsistent clips were discarded, resulting in 30,000 refined samples that maintain balance across all seven emotions.

\begin{figure*}[htbp!]
\centering
\begin{subfigure}[b]{0.30\textwidth}
    \centering
    \includegraphics[width=\textwidth]{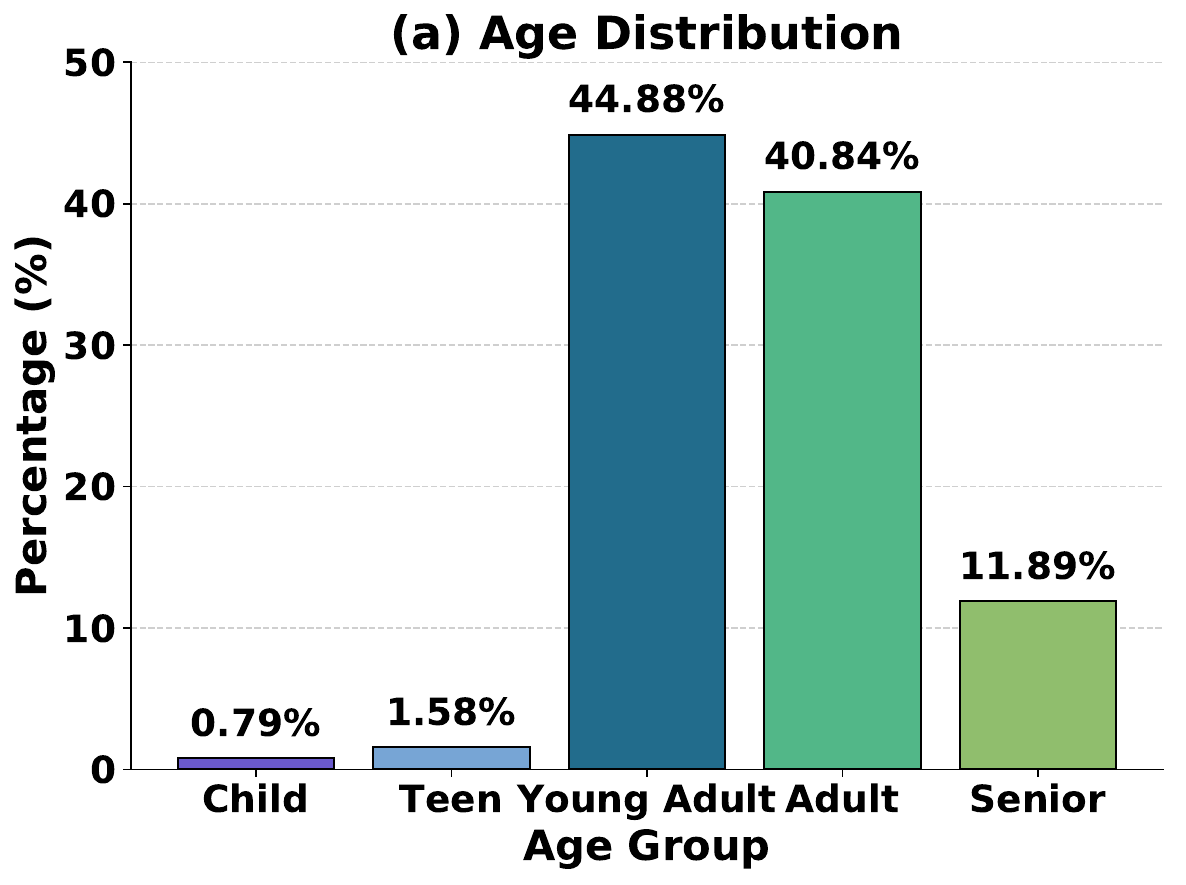}
    \label{subfig:class_dist}
\end{subfigure}
\begin{subfigure}[b]{0.30\textwidth}
    \centering
    \includegraphics[width=\textwidth]{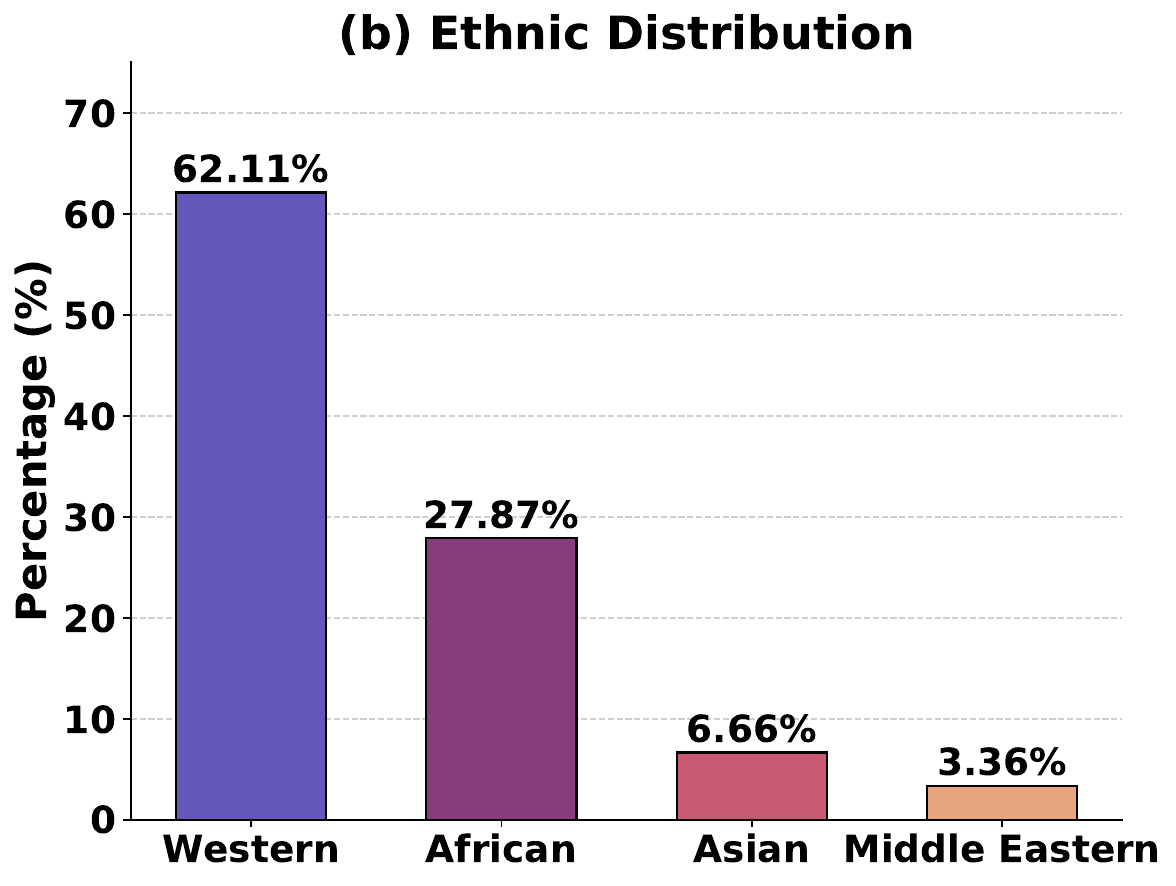}
    \label{subfig:perf_comp}
\end{subfigure}
\begin{subfigure}[b]{0.20\textwidth}
    \centering
    \includegraphics[width=\textwidth]{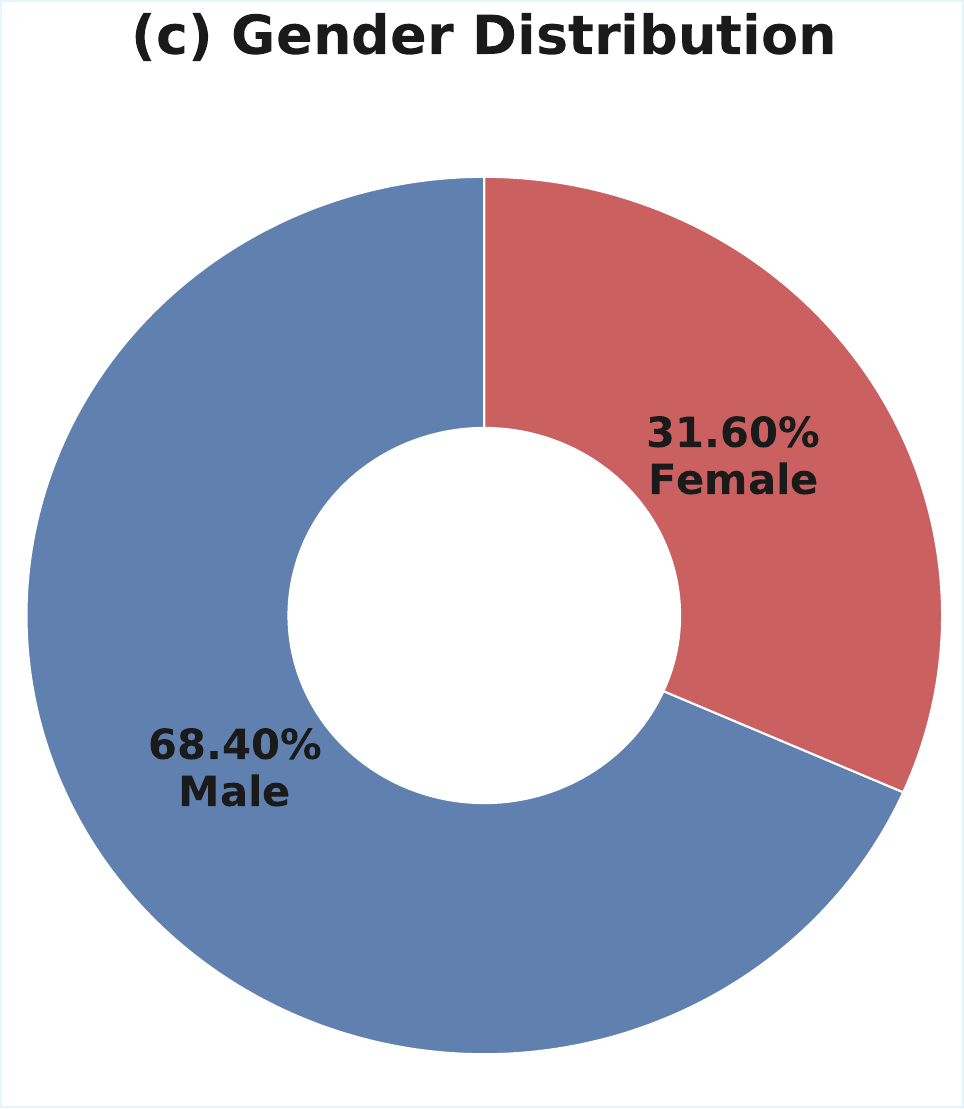}
    \label{demograhic_dist}
\end{subfigure}
\vspace{-6mm}
\caption{Demographic distribution of the SpEmoC dataset across age groups, ethnicity, and gender, illustrating its diversity and balanced population coverage.}
\label{demographic_dist}
\vspace{-5mm}
\end{figure*}

\noindent\textbf{SpEmoC Demographics:}
This section summarizes the demographic composition of the dataset (30K refined clips) across age, gender, and ethnicity. As shown in Figure \ref{demographic_dist}, the dataset spans a wide age range from children to seniors and includes both male and female participants from multiple ethnic groups. After dataset refinement, demographic attributes (age group, gender, ethnicity) were estimated using a pretrained visual demographic model and verified through manual cross-checking. Since these attributes are inferred from visual content in movie clips, they remain approximate and may be affected by factors such as makeup, lighting, or role-based appearance changes.

\noindent\textbf{Dataset Splitting Strategy:}
\label{dataset_split}
To prevent content leakage, we adopt a movie-level split assigning complete movies to train (70\%), validation (10\%), or test (20\%) sets.
This ensures no overlap in scenes, actors, or dialogues between splits, enabling robust real-world evaluation. This \textit{movie-independent split} ensures all related sequels or series episodes remain within a single set, preventing actor or scene overlap and enabling robust real-world generalization and guaranteeing an unseen test set.
The final dataset statistics and class-wise distributions are compared against MELD and CAER in Figure \ref{distribution}(a) and Table \ref{tab:class_comparision} detailing class distributions across splits. The dataset split information is provided in the \textbf{\textcolor{blue}{Supplementary Sec: 7}}.
Overall, this annotation-refinement pipeline integrates multimodal fusion, threshold-based filtering, and expert validation to produce a balanced, trustworthy dataset suitable for fair multimodal emotion recognition.

\begin{table}[htbp!]
\vspace{-5mm}
\centering
\small
\renewcommand{\arraystretch}{0.5}
\caption{Comparison of emotion class distributions across training, validation, and test splits for MELD~\cite{poria2019meld}, CAER~\cite{lee2019context}, and the proposed SpEmoC dataset. \textcolor{red}{Red} indicates underrepresented classes (e.g., Disgust, Fear), while \textcolor{blue}{Blue} denotes the overrepresented Neutral class in existing datasets.}
\small
\begin{tabular}{l|ccc|ccc|ccc}
\hline
\multirow{2}{*}{\textbf{Categories}} & \multicolumn{3}{c|}{\textbf{MELD}} & \multicolumn{3}{c|}{\textbf{CAER}} & \multicolumn{3}{c}{\textbf{ SpEmoC (Ours)}} \\
\cline{2-10}
 & Train & Val & Test & Train & Val & Test & Train & Val & Test\\
\hline
Anger   & 1109 & 153 & 345 &1136 & 162& 325 & 3980  & 392 & 1271   \\
Disgust  & \textcolor{blue}{271}  & \textcolor{blue}{22}  & \textcolor{blue}{68} &  \textcolor{blue}{500} & \textcolor{blue}{71} & \textcolor{blue}{145} & 4946  & 551  & 1298     \\
Fear     & \textcolor{blue}{268}  & \textcolor{blue}{40}  & \textcolor{blue}{50} & \textcolor{blue}{358} & \textcolor{blue}{51} & \textcolor{blue}{102} & 2378 & 226  & 729     \\
Joy      & 1743 & 163 & 402 & 1905& 272 & 544 & 2506 & 340 & 578   \\
Neutral  & \textcolor{red}{4710} & \textcolor{red}{470} & \textcolor{red}{1256} & \textcolor{red}{3202} & \textcolor{red}{457} & \textcolor{red}{915} & 2804 & 320 & 895  \\
Sadness   & 683  & 111 & 208 & 1028 & 146 & 294 & 1612  & 195  & 474   \\
Surprise & 1205 & 150 & 281 & 1093 & 157 & 315 & 3181 & 376 & 811   \\
\hline
\end{tabular}
\label{tab:class_comparision}
\vspace{-5mm}
\end{table}

\noindent\textbf{Ethical Considerations:} We prioritize copyright and responsible use in constructing SpEmoC. The dataset is derived from publicly available movies and TV series and will be released strictly under fair-use provisions for non-commercial research. Access will be regulated through an End User License Agreement (EULA), requiring researchers to apply for access and comply with clearly defined terms. To ensure transparency, the dataset repository will provide detailed documentation on usage boundaries, licensing conditions, and ethical safeguards.\\

\section{Benchmarking and Analysis of SpEmoC}
\label{results}
\vspace{-2mm}
To ensure fair evaluation of \textbf{SpEmoC}, all experiments use identical training protocols, optimization settings, and metrics across MELD, CAER, and SpEmoC. We benchmark five multimodal emotion recognition models MulT~\cite{tsai2019multimodal}, MISA~\cite{hazarika2020misa}, EmotionCLIP~\cite{zhang2023learning}, TCL-MAP~\cite{zhou2024token}, and EMOE~\cite{fang2025emoe} using consistent 70/10/20 source-level splits.\\
The evaluation covers five aspects: in-domain benchmarking, cross-dataset generalization, low-data transfer learning, neutral-class removal analysis, and modality-level transfer. Performance is reported using Weighted F1, Macro-F1, per-class F1 and imbalance gap $\Delta = \text{W-F1} - \text{Macro-F1}$. Additional implementation details are provided in the \textcolor{blue}{\textbf{\textit{Supplementary Sec: 3}}}.
\subsection{In-Domain Benchmarking on SpEmoC}
We evaluate five representative multimodal emotion recognition models on SpEmoC and compare them with two widely used benchmarks, MELD and CAER. While W-F1 reflects overall accuracy, Macro-F1 measures balanced performance across emotion classes, and $\Delta$ quantifies bias toward majority categories.
Table~\ref{SOTA_results} shows that all models achieve substantially higher Macro-F1 on SpEmoC than on MELD and CAER, with improvements exceeding 30 absolute points on average. For instance, EMOE improves from 38.19 (MELD) and 15.54 (CAER) to 70.25 on SpEmoC, while TCL-MAP increases from 45.46 and 20.99 to 73.34. Similar gains are observed for MulT, MISA, and EmotionCLIP, indicating that the improvements arise from the dataset design rather than model architecture.\\
A key observation is the dramatic reduction of the imbalance gap ($\Delta$). On MELD and CAER the gap typically ranges between 15-25 points, reflecting strong bias toward dominant classes such as \textit{Neutral}. In contrast, SpEmoC consistently produces near-zero imbalance gaps ($\Delta \leq 0.65$), indicating much more uniform performance across the seven emotion categories.
Per-class analysis further confirms this trend. Minority emotions such as \textit{Fear} and \textit{Disgust}, which often obtain near-zero F1 scores on MELD and CAER, achieve substantially higher performance on SpEmoC (typically 36-71 F1). These results demonstrate that the balanced emotion distribution. SpEmoC provide stronger supervision signals for multimodal emotion learning.
The consistent gains across architectures indicate that SpEmoC mitigates class imbalance and provides a reliable benchmark for multimodal emotion recognition.
\begin{table*}[htbp!]
\centering
\small
\renewcommand{\arraystretch}{1}
\caption{In-Domain Per-Class Performance and Imbalance Analysis Across \textbf{SpEmoC}, MELD, and CAER. Lower imbalance gap ($\Delta$) indicates better class balance. \textbf{Bold}: Best}
\label{SOTA_results}
\begin{adjustbox}{max width=.9\textwidth}
\begin{tabular}{cccccccccccc}
\hline
\textbf{Methods} & \textbf{Dataset} &
\textbf{Surprise} & \textbf{Joy} & \textbf{Fear} & \textbf{Disgust} &
\textbf{Anger} & \textbf{Neutral} & \textbf{Sadness} & \textbf{W-F1 $\uparrow$} & \textbf{Macro-F1$\uparrow$} & \textbf{$\Delta$(Gap$\downarrow$})\\
\hline

\multirow{3}{*}{MISA~\cite{hazarika2020misa}}
 & MELD  & 13.30 & 18.93 & 0.74 & 0.0 & 23.52 & 37.95 & 14.33 & 25.09 & 18.61 & 6.48\\
 & CAER  & 18.72 & 58.08 & 0.0 & 15.71 & 17.80 & 49.33 & 21.92 & 32.82 & 25.94 & 6.88\\
 & \textbf{SpEmoC} & 41.30 & 66.40 & 36.00 & 49.80 & 50.90 & 32.40 & 51.70 & 50.78 & 47.50 & \textcolor{green}{\textbf{3.28}}\\
\hline

\multirow{3}{*}{EmotionCLIP~\cite{zhang2023learning}}
 & MELD  & 26.28 & 39.72 & 0.0 & 0.0 & 28.44 & 58.63 & 17.21 & 34.59 & 24.33 & 10.26\\
 & CAER  & 11.94 & 34.88 & 0.0 & 0.0 & 21.37 & 46.52 & 13.76 & 27.45 & 18.35& 9.10 \\
 & \textbf{SpEmoC} & 52.63 & 66.93 & 50.49 & 55.82 & 51.60 & 31.76 & 48.47 & 51.30 & \textbf{49.50} & \textcolor{green}{\textbf{1.8}}\\
\hline

\multirow{3}{*}{TCL-MAP~\cite{zhou2024token}} 
& MELD  & 56.04 & 56.44 & 17.07 & 22.22 & 46.05 & 77.75 & 33.93 & 62.68 & 45.46 & 17.22 \\
 & CAER  & 13.24 & 28.73 & 10.85 & 7.51 & 23.37 & 44.79 & 12.98 & 28.26 & 20.99 & 7.27 \\
 & \textbf{SpEmoC} & 74.36 & 75.55 & 71.13 & 70.84 & 73.76 & 75.51 & 72.25 & 73.67 & \textbf{73.34 }& \textcolor{green}{\textbf{0.33}}\\
\hline

\multirow{3}{*}{EMOE~\cite{fang2025emoe}}
 & MELD  & 54.59 & 54.55 & 6.06 & 6.38 & 43.95 & 73.86 & 27.92 & 59.04 & 38.19 & 20.85\\
 & CAER  & 0.0 & 28.66 & 0.0 & 0.0 & 24.34 & 51.95 & 3.82 & 36.52 &15.54 & 20.98\\
 & \textbf{SpEmoC} & 80.94 & 84.73 & 66.42 & 66.78 & 65.85 & 61.01 & 66.05 & 70.30 & \textbf{70.25} & \textcolor{green}{\textbf{0.05}}\\

 \midrule
 
\multirow{3}{*}{MulT\cite{tsai2019multimodal}}
 & MELD  & 49.74 & 55.42 & 7.27 & 9.09 & 42.46 & 74.08 & 32.42 & 58.85 & 38.64 & 20.21\\
 & CAER  & 9.84 & 22.99 & 0.0 & 0.0 & 4.02 & 52.16 & 0.0 & 36.03 & 12.72 & 23.31\\
 & \textbf{SpEmoC} & 47.10 & 75.26 & 40.44 & 60.06 & 60.05 & 35.13 & 60.47 & 53.37 &\textbf{52.72} & \textcolor{green}{\textbf{0.65}}\\
\hline
 
\hline
\end{tabular}
\end{adjustbox}
\vspace{-4mm}
\end{table*}

\subsection{Cross-Dataset Generalization Analysis}
To measure transfer robustness, we perform direct cross-dataset evaluation each model is trained on one dataset and tested on the others without adaptation (Tables~\ref{tcl_cross_dataset}--\ref{emoe_cross_dataset}).\\
\noindent\textbf{Overall trend:}
Across TCL-MAP, EMOE, MulT, and MISA, training on SpEmoC yields more balanced class-wise performance and improved robustness to class imbalance than training on MELD or CAER. While MELD$\leftrightarrow$CAER transfers are often stronger than transfers involving SpEmoC (reflecting their closer distributional similarity and shared Neutral-dominant bias), SpEmoC-trained models degrade more gracefully across transfer directions and avoid brittle failures.

\noindent\textbf{Minority-class stability:}
The main difference appears in rare emotions. When trained on MELD/CAER, models frequently collapse on \textit{Fear} and \textit{Disgust} under transfer (near-zero F1 in several directions; e.g., MulT Table/MISA tables \ref{mult_cross_dataset} and \ref{misa_cross_dataset}). In contrast, SpEmoC-trained models retain non-trivial F1 for these classes across architectures, indicating improved minority-emotion generalization rather than overfitting to \textit{Neutral}.

\noindent\textbf{Dataset Similarity (MELD vs CAER):}
MELD and CAER show stronger mutual transfer than transfers involving SpEmoC (Tables~\ref{tcl_cross_dataset}--\ref{misa_cross_dataset}; e.g., TCL-MAP: 64.21 W-F1, EMOE: 35.61 W-F1), indicating closer distributions and Neutral-class bias compared to the more balanced SpEmoC.

\noindent\textbf{Model-Agnostic Consistency:}
Improvements are consistent across architectures (TCL-MAP, MulT, MISA, EMOE; Tables~\ref{tcl_cross_dataset}--\ref{misa_cross_dataset}), suggesting that the observed improvements are associated with SpEmoC's overall dataset design rather than model-specific factors.

\noindent\textbf{Imbalance-gap interpretation:}
Consistent with the above, MELD/CAER training produces large imbalance gaps ($\Delta=15$-$27$; Tables~\ref{tcl_cross_dataset}--\ref{misa_cross_dataset}), indicating strong bias toward the Neutral class. In contrast, SpEmoC training yields much smaller gaps across architectures-TCL-MAP ($\Delta=0.33$, Table~\ref{tcl_cross_dataset}), EMOE ($\Delta=0.34$, Table~\ref{emoe_cross_dataset}), MulT ($\Delta=0.65$, Table~\ref{mult_cross_dataset}), and MISA ($\Delta=3.28$, Table~\ref{misa_cross_dataset}) reflecting more balanced class-wise recognition.

\noindent\textbf{Connection to downstream results:}
These findings help explain why SpEmoC pretraining improves (i) fine-tuning on MELD/CAER (Table~\ref{fitetune_spemoc}) and (ii) low-data adaptation (Table~\ref{table_lowdata}), balanced supervision in SpEmoC preserves discriminative features for minority emotions, which remain useful under domain shift and limited supervision.

\begin{table*}[htbp!]
\vspace{-5mm}
\centering
\caption {Cross-dataset generalization across SpEmoC (ours), MELD, and CAER using the TCL-MAP \cite{zhou2024token} model, showing more balanced performance when trained on SpEmoC.}
\label{tcl_cross_dataset}
\vspace{-3mm}
\renewcommand{\arraystretch}{1}
\begin{adjustbox}{max width=.80\textwidth}
\begin{tabular}{lccccccccccc}
\toprule
\textbf{Train → Test} &
\textbf{Surprise} & \textbf{Joy} & \textbf{Fear} & \textbf{Disgust} & 
\textbf{Anger} & \textbf{Neutral} & \textbf{Sadness} &
\textbf{W-F1} $\uparrow$ &\textbf{Macro-F1} $\uparrow$ & \textbf{$\Delta$(Gap$\downarrow$})\\
\midrule

CAER → CAER     & 13.24 & 28.73 & 10.85 &  7.51 & 23.37 & 44.79 & 12.98 & 28.26 & 20.21 & 8.05 \\
CAER → MELD     & 19.46 & 17.44 &  5.13 &  1.14 & 16.09 & 50.71 & 13.56 & 34.62 &  17.65 & 16.97  \\
CAER → SpEmoC(ours)   & 21.14 & 34.79 &  3.49 &  5.47 & 40.94 & 25.89 & 17.07 & 24.64 &  19.83 & \textbf{4.81 } \\
\midrule

MELD → CAER     & 53.27 & 57.32 & 15.58 & 17.82 & 44.59 & 79.91 & 34.06 & 64.21 &  43.22 & 20.99  \\
MELD → MELD     & 56.04 & 56.44 & 17.07 & 22.22 & 46.05 & 77.75 & 33.93 & 62.68 & 45.46 &17.22\\
MELD → SpEmoC(ours)   & 44.83 & 53.47 &  3.92 & 18.21 & 51.28 & 34.91 & 30.77 & 36.00 & 32.17 & \textbf{3.83 }  \\
\midrule

SpEmoC(ours) → CAER   & 21.31 & 17.92 & 10.78 & 13.30 & 20.30 & 46.69 & 11.36 & 30.75 &  20.82 & 9.93   \\
SpEmoC(ours) → MELD   & 45.19 & 29.37 &  3.96 & 16.72 & 39.71 & 69.59 & 25.24 & 50.73 &  32.83 & 17.90  \\
SpEmoC → SpEmoC & 75.51 & 74.36 & 71.13 & 72.25 & 75.55 & 70.84 & 73.76 & 73.67 & 73.34 & \textbf{0.33}\\

\bottomrule
\end{tabular}
\end{adjustbox}
\vspace{-5mm}
\end{table*}

\begin{table}[H]
\vspace{-3mm}
\centering
\caption {Cross-dataset generalization across SpEmoC (ours), MELD, and CAER using the MISA~\cite{hazarika2020misa} model.}
\label{misa_cross_dataset}
\renewcommand{\arraystretch}{1}
\begin{adjustbox}{max width=.80\textwidth}
\begin{tabular}{lccccccccccc}
\toprule
\textbf{Train → Test} &
\textbf{Surprise} & \textbf{Joy} & \textbf{Fear} & \textbf{Disgust} & 
\textbf{Anger} & \textbf{Neutral} & \textbf{Sadness} &
\textbf{W-F1} $\uparrow$ &\textbf{Macro-F1} $\uparrow$ & \textbf{$\Delta$(Gap$\downarrow$})  \\
\midrule
CAER → CAER     & 18.72 & 58.08 & 0.0 & 15.71 & 17.80 & 49.33 & 21.92 & 32.82 &25.94&
6.88\\
CAER → MELD     & 19.10 & 29.45 &  0.0 & 5.47 & 20.62 & 11.63 & 4.34 & 14.86 & 14.29 & \textbf{0.57}   \\
CAER → SpEmoC   & 0.24 & 12.76 &  0.0 & 6.98 & 32.58 & 12.07 & 0.41 & 19.42 &   10.15 & 9.27 \\
\midrule

MELD → CAER     & 12.29 & 46.21 &  0.0 & 0.0 & 19.70 & 51.20 & 3.78 & 38.64 &  19.88 & 18.76  \\
MELD → MELD     & 13.30 & 18.93 & 0.74 & 0.0 & 23.52 & 37.95 & 14.33 & 25.09 & 18.61 & \textbf{6.48} \\
MELD → SpEmoC   & 15.77 & 27.39 &  0.0 & 0.0 & 34.49 & 28.65 & 0.0 & 22.65 &  15.19 & 7.46  \\

\midrule
SpEmoC → CAER   & 5.57 & 8.21 &  6.54 & 7.82 & 9.07 & 42.04 & 12.14 & 23.77 & 13.06 & 10.71   \\
SpEmoC → MELD   & 9.91 & 4.37 &  3.99 & 5.53 & 13.50 & 42.00 & 4.19 & 21.73 & 13.64 & 8.09   \\
SpEmoC → SpEmoC  & 41.30 & 66.40 & 36.00 & 49.80 & 50.90 & 32.40 & 51.70 & 50.78 & 47.50 & \textbf{3.28}\\

\bottomrule
\end{tabular}
\end{adjustbox}
\end{table}

\begin{table}[H]
\vspace{-10mm}
\centering
\caption {Cross-dataset generalization across SpEmoC (ours), MELD, and CAER using the MulT~\cite{tsai2019multimodal} model.}
\label{mult_cross_dataset}
\renewcommand{\arraystretch}{1}
\begin{adjustbox}{max width=.80\textwidth}
\begin{tabular}{lccccccccccc}
\toprule
\textbf{Train → Test} &
\textbf{Surprise} & \textbf{Joy} & \textbf{Fear} & \textbf{Disgust} & 
\textbf{Anger} & \textbf{Neutral} & \textbf{Sadness} &
\textbf{W-F1} $\uparrow$ &\textbf{Macro-F1} $\uparrow$ & \textbf{$\Delta$(Gap$\downarrow$}) \\
\midrule

CAER → CAER   & 9.84 & 22.99 &  0.0 & 0.0 & 4.02 & 52.16 & 0.0 & 36.03 & 12.72 & 23.31   \\
CAER → MELD    & 33.47 & 40.58 &  0.0 & 0.0 & 9.15 & 65.09 & 0.0 & 48.89 &  21.18 & 27.71  \\
CAER → SpEmoC   & 0.0 & 19.79 &  0.0 & 0.0 & 15.79 & 26.63 & 0.0 & 17.45 &  8.89 & \textbf{8.56}  \\
\midrule

MELD → CAER   & 16.63 & 18.50 &  0.0 & 2.50 & 16.63 & 48.55 & 19.51 & 32.61&  17.62 &14.99  \\
MELD → MELD    & 49.74 & 55.42 &  7.27 & 9.09 & 42.46 & 74.08 & 32.42 & 58.85 &  38.64 &20.21  \\
MELD → SpEmoC   & 9.77 & 48.31 &  0.55 & 6.74 & 35.23 & 30.62 & 25.35 & 25.97 &  20.96 &\textbf{ 5.01 } \\
\midrule

SpEmoC → CAER    & 20.14 & 27.95 &  7.37 & 11.28 & 14.69 & 19.65 & 18.49 & 19.19 &  17.08 & 2.11  \\
SpEmoC → MELD  & 31.70 & 46.48 &  8.26 & 11.36 & 31.54 & 26.25 & 29.35 & 30.73 &  26.56 & 4.17  \\
SpEmoC → SpEmoC    & 47.10 & 75.26 &  40.44 & 60.06 & 60.05 & 35.13 & 60.47 & 53.37 &  52.72 & \textbf{0.65}  \\
\bottomrule
\end{tabular}
\end{adjustbox}
\vspace{-3mm}
\end{table}

\begin{table}[H]
\centering
\vspace{-8mm}
\caption {Cross-dataset generalization across SpEmoC (ours), MELD, and CAER using the EMOE~\cite{fang2025emoe} model.}
\label{emoe_cross_dataset}

\renewcommand{\arraystretch}{1}
\begin{adjustbox}{max width=.80\textwidth}
\begin{tabular}{lccccccccccc}
\toprule
\textbf{Train → Test} &
\textbf{Surprise} & \textbf{Joy} & \textbf{Fear} & \textbf{Disgust} & 
\textbf{Anger} & \textbf{Neutral} & \textbf{Sadness} &
\textbf{W-F1} $\uparrow$ &\textbf{Macro-F1} $\uparrow$ & \textbf{$\Delta$(Gap$\downarrow$}) \\
\midrule

CAER → CAER & 0.0 & 28.66 & 0.0 & 0.0 & 24.34 & 51.95 & 3.82 & 36.52 &15.64
& 20.88\\
CAER → MELD    & 0.71 & 37.34 &  0.0 & 0.0 & 28.57 & 67.31 & 0.93 & 49.92 & 19.27 & 30.65    \\
CAER → SpEmoC   & 0.25 & 10.09 &  0.0 & 0.0 & 32.61 & 27.45 & 0.82 & 19.68 &  10.17 & \textbf{9.51}  \\
\midrule

MELD → CAER   & 17.75 & 19.14 &  0.0 & 1.32 & 14.18 & 52.00 & 7.63 & 35.61 &  15.15 & 20.46  \\
MELD → MELD & 54.59 & 54.55 & 6.06 & 6.38 & 43.95 & 73.86 & 27.92 & 59.04 & 38.19 &20.85 \\
MELD → SpEmoC   & 38.12 & 52.45 &  1.61 & 23.45 & 44.08 & 34.73 & 28.17 & 34.62 &  31.01 & \textbf{3.61}  \\
\midrule

SpEmoC → CAER    & 20.69 & 24.18 &  5.88 & 9.57 & 18.61 & 45.88 & 11.11 & 30.83 &  19.56 &11.27  \\
SpEmoC → MELD  & 41.56 & 26.30 &  6.06 & 12.42 & 34.33 & 65.50 & 23.61 & 47.74 & 29.97 & 17.77   \\
SpEmoC → SpEmoC & 80.96 & 84.55 & 68.59 & 71.49 & 71.44 & 62.25 & 66.05 & 71.85 & 72.19 & \textbf{0.34}\\
\bottomrule
\end{tabular}
\end{adjustbox}
\vspace{-5mm}
\end{table}

\subsection{Comprehensive Dataset Analysis}
\label{sec:ablation}
We conduct a series of comprehensive analyses to evaluate 
the robustness, generalization, and practical utility of the proposed SpEmoC dataset. Rather than focusing on architectural novelty, we systematically analyze the impact of dataset design choices across datasets, modalities, data availability, and class balance using standard multimodal emotion recognition frameworks. All experiments are conducted using the EMOE~\cite{fang2025emoe} model under identical training settings.\\
\noindent\textbf{Low-Data Transfer Learning Performance:}
We evaluate fine-tuning on MELD and CAER using 10\%, 30\%, and 50\% of training data (Table~\ref{table_lowdata}). SpEmoC-pretrained models consistently outperform non-pretrained baselines, with the largest gains in the extreme low-data regime (10\%). This indicates that SpEmoC provides transferable emotional representations that remain effective under limited supervision.
\begin{table}[htbp!]
\centering
\small
\caption{Low-data transfer learning results when fine-tuning on MELD and CAER using only 10\%, 30\%, and 50\% of the training data. SpEmoC pretraining consistently improves performance across all data regimes.}
\label{table_lowdata}
\vspace{-3mm}
\renewcommand{\arraystretch}{0.6}
\begin{adjustbox}{max width=.8\textwidth}
\begin{tabular}{lcc @{\hspace{15pt}}c  @{\hspace{15pt}}c}
\toprule
\textbf{Training Split} &
\textbf{Baseline } &
\textbf{+SpEmoC Pretraining (W-F1 $\uparrow$ )} & \textbf{Macro-F1$\uparrow$} & \textbf{$\Delta$(Gap$\downarrow$})\\
\midrule

10\%  & \textcolor{red}{57.13} / \textcolor{green}{31.40} & \textbf{\textcolor{red}{59.16}} / \textbf{\textcolor{green}{34.70}} & \textcolor{red}{34 .60}/\textcolor{green}{15.40} & \textcolor{red}{24.46}/\textcolor{green}{19.30} \\

30\%  & \textcolor{red}{61.00} / \textcolor{green}{35.27} & \textbf{\textcolor{red}{62.40}} / \textbf{\textcolor{green}{35.34}} & \textcolor{red}{35.77}/\textcolor{green}{14.48} & \textcolor{red}{23.64}/\textcolor{green}{20.86}\\

50\%  & \textcolor{red}{59.73} / \textcolor{green}{34.89} & \textbf{\textcolor{red}{61.50}} / \textbf{\textcolor{green}{35.61}} & \textcolor{red}{38.04}/\textcolor{green}{11.66} & \textcolor{red}{22.99}/\textcolor{green}{23.95} \\
\bottomrule
\end{tabular}
\end{adjustbox}
\vspace{-5mm}
\end{table}

\noindent\textbf{Cross-dataset transfer learning:}
Table~\ref{fitetune_spemoc} shows that SpEmoC pretraining consistently improves cross-dataset performance. 
On MELD, minority-class recognition improves substantially, particularly for \textit{Fear} (6.06 $\rightarrow$ 16.33) and \textit{Disgust} (6.38 $\rightarrow$ 17.65), increasing W-F1 from 59.04 to 63.26 (+4.22). 
On CAER, gains in \textit{Surprise}, \textit{Disgust}, and \textit{Sadness} raise W-F1 from 36.52 to 38.16 (+1.64). 
Reverse transfer experiments further show consistent improvements when SpEmoC is finetuned from MELD or CAER initialization, confirming the robustness of cross-dataset representations.\\
Further analyses on backbone variants and neutral-class removal are provided in the \textbf{\textcolor{blue}{Supplementary Sec: 1 and 2}}.

\begin{table}[H]
\vspace{-5mm}
\centering
\caption{Class-wise and W-F1 improvements after SpEmoC pretraining on MELD and CAER. $\Delta$ denotes relative gain.}
\vspace{-2mm}
\renewcommand{\arraystretch}{1}
\label{fitetune_spemoc}
\begin{adjustbox}{max width=\textwidth}
\begin{tabular}{lcccccccccc}
\toprule
\textbf{Dataset} &
\textbf{Surprise} & \textbf{Joy} & \textbf{Fear} & \textbf{Disgust} &
\textbf{Anger} & \textbf{Neutral} & \textbf{Sadness} & \textbf{Micro-F1 $\uparrow$} &\textbf{W-F1 $\uparrow$} &
\textbf{$\Delta$ (gain)}\\
\midrule

\textbf{MELD Training} & 54.59 & 54.55 & 6.06 & 6.38 & 43.95 & 73.86 & 27.92 & 38.19 & 59.04 \\

\rowcolor{lightblue}\textbf{MELD Finetuning (+SpEmoC Pretraining)} &
\textbf{58.22} & \textbf{55.20} & \textbf{16.33} & \textbf{17.65} & 41.68 & \textbf{78.39} & 21.90& \textbf{41.34} & \textbf{63.26} & + 4.22\\

\midrule

\textbf{SpEmoC Training}  & 80.94 & 84.73 & 66.42 & 66.78 & 65.85 & 61.01 & 66.05 &70.25 & 70.30 \\
\rowcolor{lightblue}\textbf{SpEmoC Finetuning (+MELD Pretraining)} &
80.65 & 85.08 & 68.11 & 71.32 & 70.51 & 61.93 & 66.73 & 71.16 & 71.47 & +1.17 \\
\midrule
\textbf{CAER Training} & 0.0 & 28.66 & 0.0 & 0.0 & 24.34 & 51.95 & 3.82 & 15.64& 36.52 \\

\rowcolor{lightblue}\textbf{CAER Finetuning (+SpEmoC Pretraining)} &
\textbf{13.92} & 26.98 & 0.0 & \textbf{2.3} & 23.35 & \textbf{52.47} & \textbf{10.70} & \textbf{19.81} & \textbf{38.16} & +1.64\\
\midrule

\textbf{SpEmoC Training}  & 80.94 & 84.73 & 66.42 & 66.78 & 65.85 & 61.01 & 66.05 &70.25 & 70.30 \\
\rowcolor{lightblue}\textbf{SpEmoC Finetuning (+CAER Pretraining)} & 79.92
 & 84.08 & 68.30 & 71.41 & 72.41 & 62.55 & 66.67 & 71.34 &  71.84 & +1.54 \\
\bottomrule
\end{tabular}
\end{adjustbox}
\vspace{-5mm}
\end{table}
\section{Conclusion and Future Work}
\vspace{-2mm}
We introduced \textbf{SpEmoC}, a large-scale multimodal emotion recognition benchmark containing 30,000 high-quality clips curated from 306,544 speaking segments across 3,100 English-language movies and TV series. Constructed through a hybrid pipeline combining pretrained models (DistilRoBERTa, Wav2Vec 2.0, YOLOv8), systematic filtering, and human validation, SpEmoC provides synchronized text, audio, and visual modalities with a near-balanced distribution across seven emotions. By organizing samples around short dialogue-centered segments and enforcing strict movie/series-level splits, the dataset mitigates class imbalance, improves multimodal alignment, and prevents content leakage across splits. 
Extensive experiments across multiple multimodal frameworks demonstrate improved cross-dataset generalization to MELD and CAER, stronger recognition of minority emotions such as fear and disgust, and substantially reduced imbalance gaps. Furthermore, SpEmoC pretraining consistently improves performance in both fine-tuning and low-data settings, highlighting the importance of balanced dataset design for robust multimodal emotion recognition. Future work includes extending SpEmoC with fine-grained, continuous, multi-label, intensity-based, and LLM-assisted affect annotations. We also plan to improve cultural and demographic diversity, multilingual coverage, metadata richness, and spontaneous conversational content to support fairer and more realistic emotion recognition.


%
%
\clearpage
\bibliographystyle{splncs04}
\bibliography{main}
\end{document}